\documentclass{article}
\usepackage{spconf,amsmath,amssymb,graphicx,hyperref}

\newcommand{\figref}[1]{Fig.~\ref{fig:#1}}
\newcommand{\tabref}[1]{Table~\ref{tab:#1}}
\newcommand{\secref}[1]{Section~\ref{sec:#1}}

\newcommand{\ablationquant}{
\begin{table}[t!]
    \centering
    \begin{tabular}{l|cccc}
         \hline
         & \multicolumn{4}{c}{$\Delta\text{E} 2000$ ($\downarrow$)} \\
         \cline{2-5}
         \multicolumn{1}{c|}{Model} & Mean & Q1 & Q2 & Q3 \\
         \hline
         3D LUTs \cite{zeng2020learning} & 4.39 & 2.67 & 3.74 & 5.57 \\
         3D LUTs + Easy positive & 4.04 & 2.13 & 3.41 & 5.22 \\
         3D LUTs + Hard positive & 3.68 & 1.67 & 2.82 & 4.84 \\
         3D LUTs + LAB & 4.06 & 1.87 & 3.40 & 5.08 \\
         WB LUTs (Ours) & \textbf{3.27} & \textbf{1.13} & \textbf{2.43} & \textbf{4.43} \\
         \hline
    \end{tabular}
    \caption{Ablation study comparing base 3D LUTs trained with the different contrastive learning strategies and color spaces, evaluated on Rendered WB Set1 validation split.}
    \label{tab:ablation-quant}
\end{table}
}

\newcommand{\quant}{
\begin{table*}[t!]
\centering
\resizebox{0.9\linewidth}{!}{

\begin{tabular}{l|cccc|cccc|c|c|c}
\hline
 & \multicolumn{4}{c|}{$\Delta\text{E} 2000 (\downarrow) $}  & \multicolumn{4}{c|}{MAE ($\downarrow$)} & Time & Params & Size \\
\cline{2-9}
\multicolumn{1}{c|}{Method} & Mean & Q1 & Q2 & Q3 & Mean & Q1 & Q2 & Q3 & (ms) & (K) & (MB) \\ \hline
\multicolumn{12}{c}{Rendered WB dataset Set2} \\ \hline
KNNWB \cite{Afifi_Price_Cohen_Brown_2019} & 5.60 & 3.43 & 4.90 & 7.06 & 4.48$^\circ$ & 2.26$^\circ$ & 3.64$^\circ$ & 5.95$^\circ$ & \underline{820} & 6108 & 21.8 \\
MixedWB \cite{Afifi_Brubaker_Brown_2022} & 6.05 & 3.45 & 4.92 & 7.20 & 4.92$^\circ$ & 2.69$^\circ$ & 4.10$^\circ$ & 6.37$^\circ$ & 1550 & \underline{1313} & \underline{5.09} \\
DeepWB  \cite{Afifi_Brown_2020} & 4.90 & \textbf{3.13} & 4.35 & 6.08 & 3.75$^\circ$ & 2.02$^\circ$ & 3.08$^\circ$ & 4.72$^\circ$ & 1090 & 4366 & 16.7 \\
WBFlow \cite{Li_Kang_Zhang_Ming_2023} & \textbf{4.64} & \underline{3.16} & \textbf{4.07} & \underline{5.56} & \underline{3.51$^\circ$} & \textbf{1.93$^\circ$} & \underline{2.92} & \underline{4.47$^\circ$} & 1080 & 7570 & 30.6 \\
WB LUTs (Ours) & \underline{4.71} & 3.26 & \underline{4.17} & \textbf{5.47} & \textbf{3.31$^\circ$} & \underline{1.95$^\circ$} & \textbf{2.77$^\circ$} & \textbf{4.23$^\circ$} & \textbf{3.7} & \textbf{593} &  \textbf{2.40} \\
\hline
\multicolumn{12}{c}{Rendered Cube dataset} \\ \hline
KNNWB \cite{Afifi_Price_Cohen_Brown_2019} &  5.68 & 3.22 & 4.61 & 6.70 & 4.12$^\circ$ & 1.96$^\circ$ & 3.17$^\circ$ & 5.04$^\circ$ & \underline{810} & 6108 & 21.8 \\
MixedWB \cite{Afifi_Brubaker_Brown_2022} & 5.03 & \textbf{2.07} & \textbf{3.12} & 7.19 & 4.20$^\circ$ & \textbf{1.39$^\circ$} & \textbf{2.18$^\circ$} & 5.54$^\circ$ & 1520 & \underline{1313} & \underline{5.09} \\
DeepWB  \cite{Afifi_Brown_2020} & 4.59 & 2.68 & 3.81 & 5.53 & 3.45$^\circ$ & \underline{1.87$^\circ$} & 2.82$^\circ$ & 4.26$^\circ$ & 1080 & 4366 & 16.7 \\
WBFlow \cite{Li_Kang_Zhang_Ming_2023} & \textbf{4.28} & \underline{2.71} & 3.77 & \underline{5.21} & \underline{3.34$^\circ$} & 1.94$^\circ$ & 2.82$^\circ$ & \underline{4.11$^\circ$} & 1070 & 7570 & 30.6 \\
WB LUTs (Ours) & \underline{4.33} & 2.84 & \underline{3.71} & \textbf{5.02} & \textbf{3.28$^\circ$} & 1.88$^\circ$ & \underline{2.62$^\circ$} & \textbf{3.91$^\circ$} & \textbf{3.7} & \textbf{593} &  \textbf{2.40} \\
\hline 
\end{tabular}%
}
\caption{Quantitative comparison of our method against state-of-the-art WB correction models on Rendered WB dataset Set2 \cite{Afifi_Price_Cohen_Brown_2019} and Rendered Cube dataset \cite{banic2017unsupervised}. \textbf{Best results} in each category are in bold and \underline{second-best} results are underlined.}
\label{tab:quant}
\end{table*}
}
\newcommand{\demo}{
\begin{figure}
    \centering
    \includegraphics[width=1\linewidth]{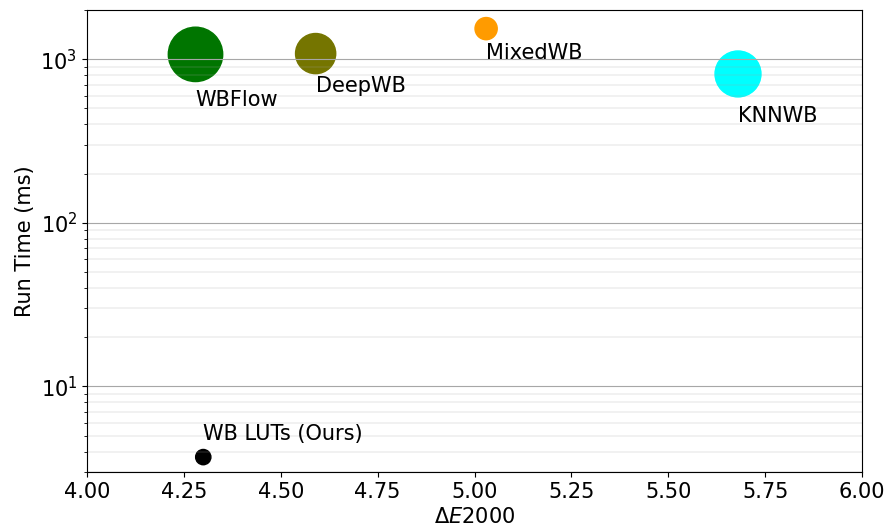}
    \caption{Comparison of performance and WB correction quality of our method with the state-of-the-art, on the Rendered Cube dataset \cite{banic2017unsupervised}. Performance is gauged both by the run time, on the vertical axis, and the model memory, which is proportional to the size of the marker for each model. }
    \label{fig:demo}
\end{figure}
}

\newcommand{\lut}{
\begin{figure}
    \centering
    \includegraphics[width=1\linewidth]{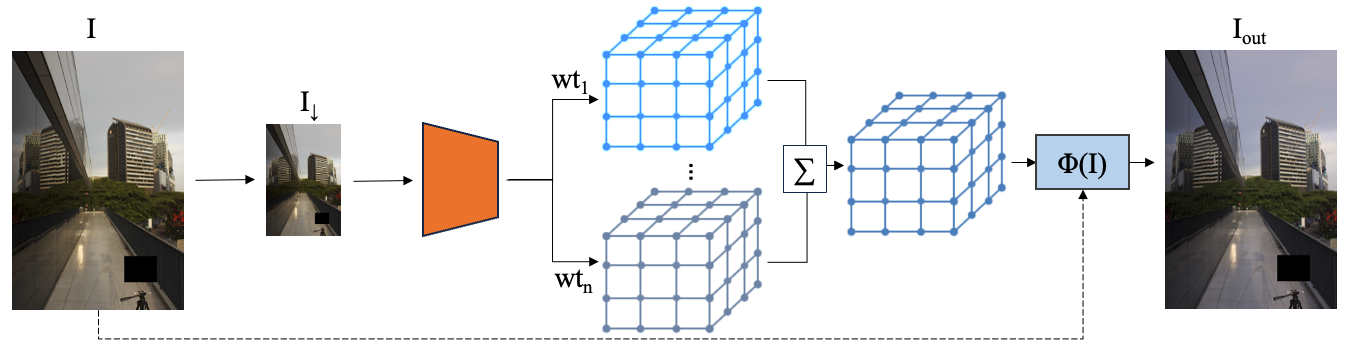}
    \caption{Downsampled input $I_{\downarrow}$ is fed to the scene classifier to generate weights for basis LUT fusion. The adaptive LUT is used to correct the high resolution image.}
    \label{fig:adaptivelut}
\end{figure}
}

\newcommand{\triplet}{
\begin{figure*}
    \centering
    \includegraphics[width=\linewidth]{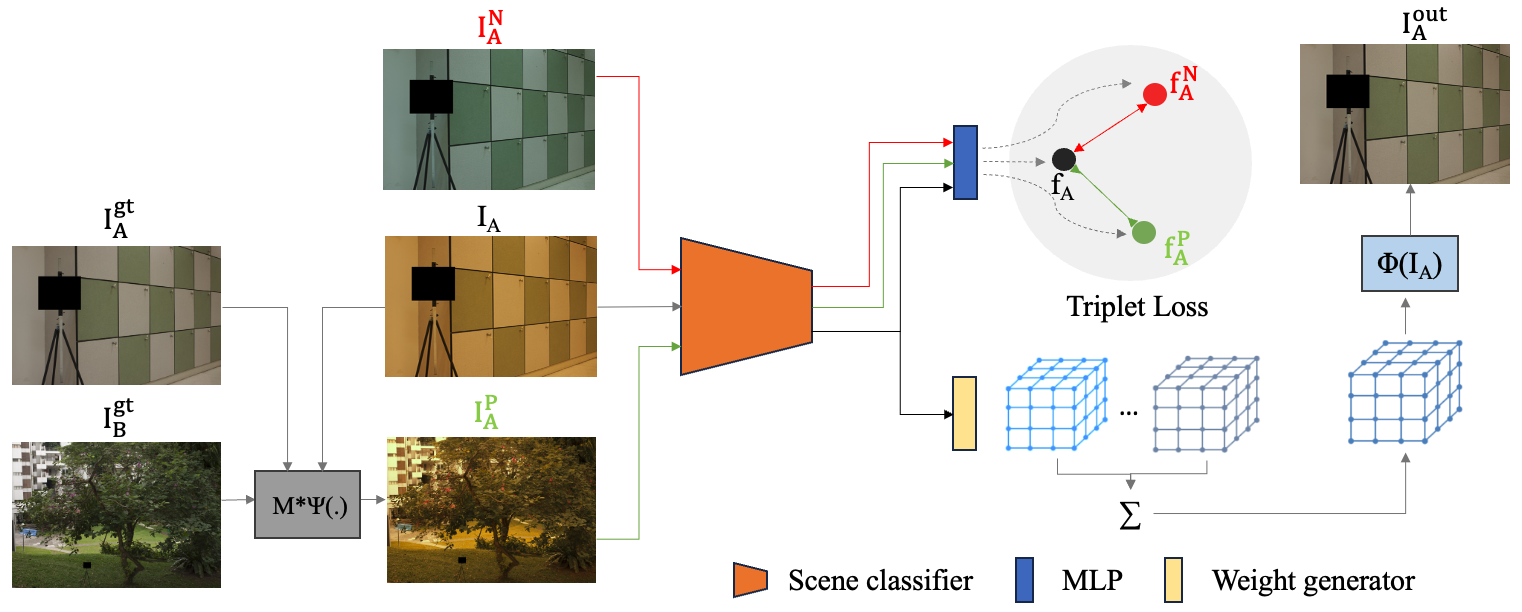}
    \caption{\textbf{Overview of proposed contrastive learning approach for LUTs.} For an image $I_A$ and its ground truth $I^\text{gt}_A$, we select an image of the same scene rendered with a different color temperature as negative sample $I^N_A$. Ground truth image of a different scene $I^\text{gt}_B$ is color mapped to generate a hard positive sample $I^{P}_A := M^\star\psi(I^\text{gt}_B)$. These hard positive and negative samples are used in the contrastive learning framework supervised with triplet loss.}
    \label{fig:arch}
\end{figure*}
}

\newcommand{\colmap}{
\begin{figure}
    \centering
    \includegraphics[width=0.8\linewidth]{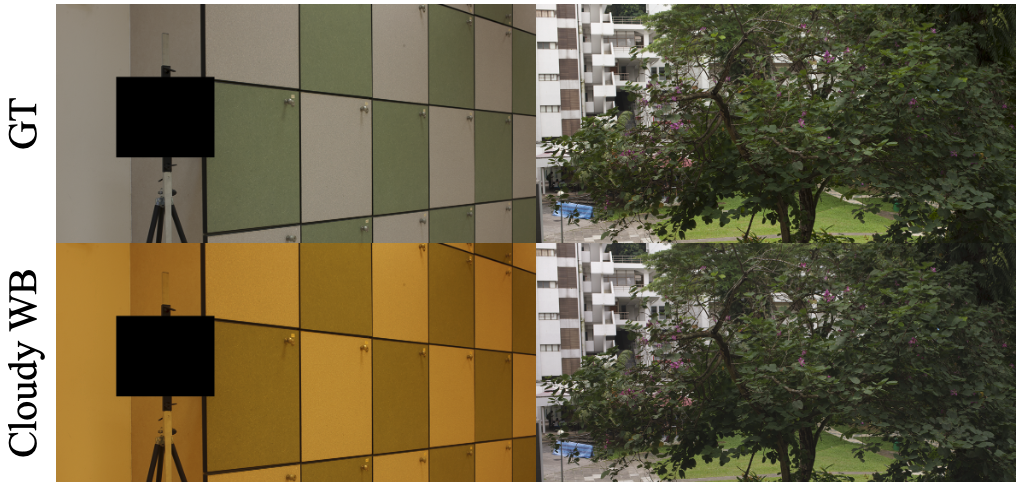}
    \caption{%Figure demonstrating the need for generating hard positive samples. 
    Two images, both rendered with the Cloudy WB setting, illustrate how the same white balance rendering can result in different color mappings for different scenes.}
    \label{fig:hard-positive}
\end{figure}
}

\newcommand{\ablationqual}{
\begin{figure}[t!]
    \centering
    \includegraphics[width=1\linewidth]{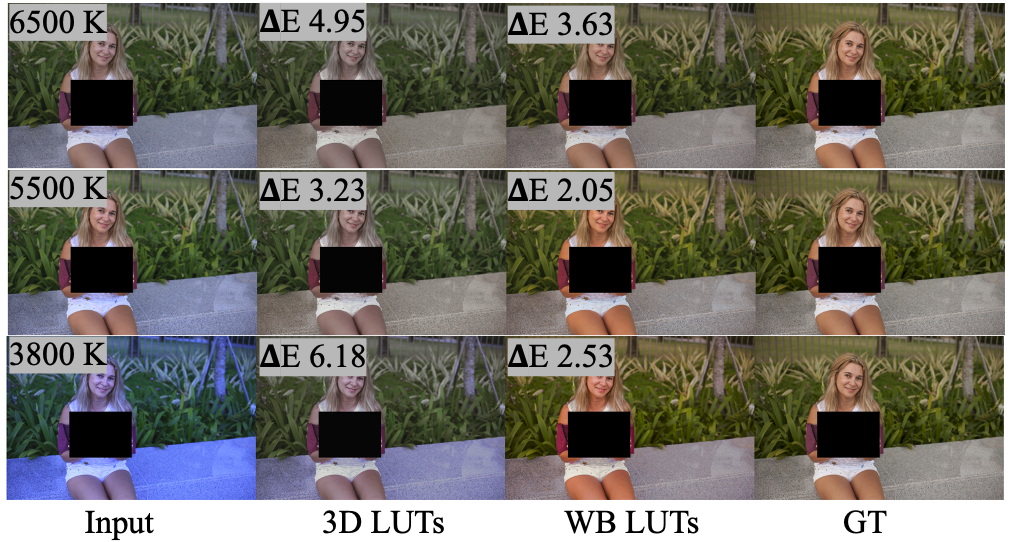}
    \caption{Results from base 3D LUTs vs WB LUTs, trained with the proposed contrastive learning framework, with $\Delta E 2000$ for each output overlaid onto the image.}
    \label{fig:ablation-qual}
\end{figure}
}

\newcommand{\qual}{
\begin{figure*}[t!]
    \centering
    \includegraphics[width=1\linewidth]{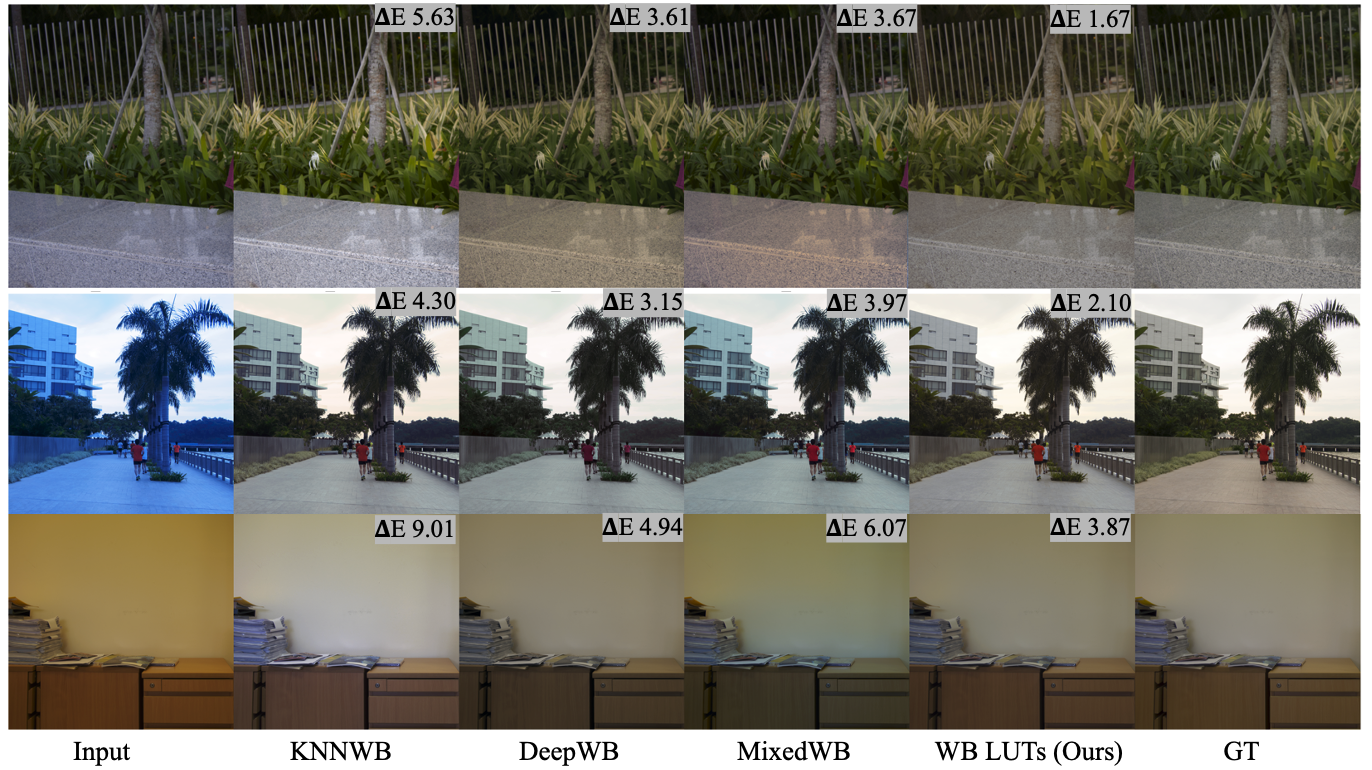}
    \caption{Qualitative comparison of our method against KNNWB \cite{Afifi_Price_Cohen_Brown_2019}, DeepWB \cite{Afifi_Brown_2020}, and MixedWB \cite{Afifi_Brubaker_Brown_2022} on images from Rendered WB Set2 \cite{Afifi_Price_Cohen_Brown_2019} and Rendered Cube \cite{banic2017unsupervised} datasets.}
    \label{fig:results}
\end{figure*}
}

\title{WB LUTs: Contrastive Learning for White Balancing Lookup Tables}

\twoauthors{Sai Kumar Reddy Manne}{Roux Institute\\ Northeastern University}{Michael Wan}{Institute for Experiential AI\\Northeastern University}
\begin{document}
%\ninept
%
\maketitle
\begin{abstract}
Automatic white balancing (AWB), one of the first steps in an integrated signal processing (ISP) pipeline, aims to correct the color cast induced by the scene illuminant. An incorrect white balance (WB) setting or AWB failure can lead to an undesired blue or red tint in the rendered sRGB image. To address this, recent methods pose the post-capture WB correction problem as an image-to-image translation task and train deep neural networks to learn the necessary color adjustments at a lower resolution. These low resolution outputs are post-processed to generate high resolution WB corrected images, forming a bottleneck in the end-to-end run time. In this paper we present a 3D Lookup Table (LUT) based WB correction model called \textbf{WB LUTs} that can generate high resolution outputs in real time. We introduce a contrastive learning framework with a novel hard sample mining strategy, which improves the WB correction quality of baseline 3D LUTs by $25.5\%$. Experimental results demonstrate that the proposed WB LUTs perform competitively against state-of-the-art models on two benchmark datasets while being $300\times$ faster using $12.7\times$ less memory. Our model and code are available at \url{https://github.com/skrmanne/3DLUT_sRGB_WB}\footnote{The authors acknowledge Bruce Maxwell for helpful discussions during the problem formulation.}.
\end{abstract}

\section{Introduction}
\label{sec:intro}
Human visual perception adaptively discerns scene colors as constant, even under different lighting conditions. Since cameras do not inherently perform this adaptation, a color balancing process is needed to transform a captured image to conform with viewer expectations. Automatic white balancing (AWB) is an important aspect of any integrated signal processing (ISP) pipeline that renders images with perceptually true object colors. However, when  WB settings are inaccurate, images are rendered with undesired color tints. Conventional color constancy methods \cite{Barron_2015, lo2021clcc}, which are based on a linear relationship between illuminant color and image intensity in the linear domain, are ineffective for post-capture white balance correction due to the non-linear operations performed in the ISP to render an sRGB image. To this end, several approaches have been proposed for WB correction based on $k$-nearest neighbors (KNN) clustering, image-to-image translation, and polynomial fitting. These models often operate on downsampled inputs to avoid computationally expensive full resolution processing. Subsequent post-processing steps, such as edge-aware upsampling \cite{Afifi_Brubaker_Brown_2022} and color mapping \cite{Afifi_Brown_2020} are employed to generate high resolution images. While these models demonstrate efficiency with low resolution inputs, the need for post-processing creates a bottleneck for real time processing, as illustrated in \figref{demo}.

\demo

By contrast, certain imaging processing pipelines use efficient 3D lookup tables (3D LUTs) to apply various finishing styles directly on the high resolution images with high efficiency. However, traditional 3D LUTs are manually tuned for each rendering style and lack generalizability. More recently, researchers have proposed methods to adaptively learn 3D LUTs for image retouching, achieving state-of-the-art quality with real time performance. Zeng \textit{et al.} \cite{zeng2020learning} propose to use a set of basis LUTs and a scene classifier to adaptively regulate each LUT's contribution to an image, creating scene-adaptive LUTs for image editing. However, unlike the image retouching task, WB correction requires different levels of color correction based on the scene and the illuminant. Hence, 3D LUTs without any direct supervision for the scene classifier struggle to generate rich illumination-oriented representations from the image. 

Khosla \textit{et al.} \cite{khosla2020supervised} proposed a supervised contrastive loss to learn representations that cluster together for similar images while pushing representations from dissimilar images farther. Motivated by the success of contrastive learning in other low-level vision tasks \cite{lo2021clcc}, we propose a contrastive learning scheme for LUTs with a novel hard sample mining strategy. The proposed contrastive learning framework encourages the model to learn illumination-oriented, scene-agnostic representations, thereby improving the WB correction efficacy. Our key contributions are as follows:
\begin{itemize}
    \item We propose \textbf{WB LUTs} for sRGB white balance correction as an alternative to image-to-image translation models. To the best of our knowledge, we are the first to use LUTs for white balance correction.
    \item We introduce a contrastive learning framework for LUTs with a novel hard sample mining strategy to learn scene-agnostic features and show $25.5\%$ improvement over baseline 3D LUTs.
    \item We achieve near state-of-the-art quality on WB correction with $300\times$ speed and $12.7\times$ smaller memory usage compared to current models.
\end{itemize}

%-------------------------------------------------------------------------

\section{Related Work}
\label{sec:related-work}
\subsection{WB Correction for sRGB Images}
Post-capture white balance correction was first proposed by Afifi \textit{et al.} \cite{Afifi_Price_Cohen_Brown_2019} along with a dataset of 65,000 images rendered with incorrect WB setting and corresponding ground truth images. For a given image, a $k$-nearest neighbours (KNN) based approach estimates a color mapping function to correct the color cast in the image. Following this work, Afifi and Brown \cite{Afifi_Brown_2020} propose a deep neural network to estimate a white balanced image along with two auxiliary outputs at tungsten and shade WB settings corresponding to 2800 K and 7500 K color temperatures respectively. The additional color temperature temperatures allow users to create images with different WB appearance through weighted fusion. In MixedWB \cite{Afifi_Brubaker_Brown_2022}, the authors improve the model's capability to handle mixed illumination settings for white balancing. By improving the ISP pipeline, they generate images at different fixed color temperatures. These images are then fed to a convolutional neural network (CNN) to estimate soft weighing masks for the weighted fusion of images at different color temperatures. The low resolution masks are upsampled using a bilateral solver to create the output at full resolution. More recently, Li \textit{et al.} \cite{Li_Kang_Zhang_Ming_2023} address the instability issue of current models and propose a stable WB model that relies on low-frequency color temperature-insensitive features to generate consistent outputs for different WB settings. Li \textit{et al.} \cite{li2023wbflow} improve the generalizability of their WB correction model using a few-shot model to learn camera-aligned pseudo raw features, reducing the errors for images from camera models not seen during the training.

\subsection{3D Lookup Tables (3D LUTs)}
Lookup tables are often used in image editing to render images with different finishing styles and are manually tuned for a specific preset style. Zeng \textit{et al.} \cite{zeng2020learning} propose image adaptive 3D LUTs for the task of image retouching and demonstrate superior quality with real time inference speeds for high resolution (4K) images. Subsequent work \cite{zhang2022dualbln, wang2021real} builds upon this initial research, enhancing the aesthetic quality of images through increased network complexity.  LUTs are rapidly being adapted to various computer vision tasks such as super resolution \cite{jo2021practical}, segmentation \cite{9345772}, image fusion \cite{Jiang_2023_ICCV}, network quantization \cite{Wang_2022_CVPR} etc. Motivated by the generalizability to several tasks, superior quality, and fast inference in image retouching, we propose a LUT-based approach for sRGB white balance correction.

%-------------------------------------------------------------------------

\section {Proposed Method}
%In this section we present details on image-adaptive LUTs, contrastive learning framework for WB LUTs, and the proposed hard sample mining strategy. 

\subsection{Revisiting 3D LUTs}
A 3D LUT is a convenient way to describe global functions for image processing. By definition, a 3D LUT is described by a 3-dimensional array $\phi$ with size $(M, M, M)$, visualized as a 3D cubic lattice. In our paper, we work with 8-bit images with corresponding color channels of 256 dimensions, and set $M=33$. For a pixel $(x,y,z)$ with intensities that each happen to be divisible by 8, the 3D LUT can be applied as a transformation $\Phi$ to $(x,y,z)$ simply by ``looking up'' the value of the scaled index in the array $\phi$: \[\Phi(x,y,z):=\phi_{(x/8,y/8,z/8)}\] For the general (non-divisible) pixel $(x,y,z)$, the value of $\Phi(x,y,z)$ is determined by trilinear interpolation of $\phi$ over the eight nearest integral divisible vertices. The 3D LUT $\phi$ can be applied to an image by applying it to each pixel uniformly. In practice, this construction encapsulates a wide range of global transformations---which are continuous with respect to pixel intensity and can be applied quickly---in a relatively small function space ($33^3=35,937$ dimensions).

Zeng \textit{et al.} \cite{zeng2020learning} proposed adaptive 3D LUTs learned from large datasets for image enhancement. A scene classifier network and $N$ basis LUTs are trained collectively to learn image-adaptive LUTs. The classifier weights are used to fuse the basis LUTs into a single image-adaptive LUT for a given image as follows:

\begin{equation}
    \phi = \sum_{n=1}^{N} w_{n}\phi^{n}
\end{equation}
where $\phi^{1..n}$ are the basis LUTs, $w_{1..n}$ are the estimated weights from the classifier, and $\phi$ is the fused LUT.

This method of training LUTs and the classifier benefits from optimal collaboration between both the image and LUT features and results in a lightweight model for image editing. Since LUTs are resolution agnostic, training and inference can be done on a downsampled image to retrieve an adaptive LUT for the task, see \figref{adaptivelut}. This adaptive LUT can be used to generate a high resolution output without loss in quality for global image retouching tasks. Since white balancing is a global color correction problem, 3D LUTs offer an attractive alternative to traditional image-to-image translation CNNs.
\lut
\subsection{Contrastive learning for WB LUTs}
\label{sec:contrastive-learning}
Following 3D LUTs \cite{zeng2020learning}, we use a scene classifier network to produce scene adaptive weights for the basis LUTs. Initial tests showed this approach to be very effective overall, but images with extreme starting color temperatures yielded images with residual color tints. We hypothesize that the lack of direct supervision for the scene classifier, resulting in scene-oriented features, is the primary cause for the degradation. In a similar task of color constancy, contrastive learning has shown significant improvements over the baseline model by generating rich illumination-oriented representations \cite{lo2021clcc}. Hence, we utilize contrastive learning as supervision for the scene classifier to learn rich illumination-oriented, scene-agnostic representations. Contrastive learning typically proceeds by comparing an \textit{anchor} sample to a similar \textit{positive} sample and a dissimilar \textit{negative} sample, with the desired contraction or separation enforced by a special loss function. In the context of white balance correction, images that require similar WB correction are considered similar even when the underlying scene is different, and conversely, images of the same scene are considered dissimilar if the WB correction required is different.
%images with casts requiring similar WB correction are considered similar, even if their underlying scenes are different, and conversely, images with similar or even the same scene but requiring different WB corrections are considered dissimilar. 

The effectiveness of contrastive learning hinges on challenging samples that are harder to distinguish. Hence, Lo \textit{et al.} \cite{lo2021clcc} propose data augmentation techniques to create \textit{hard positive} and \textit{hard negative samples}. These samples are generated by creating images with same scene content and different illuminants, as well as different scene content with same illuminants. However, such data augmentation is not possible in the post-capture WB correction task due to the lack of ground truth illumination information and raw images. Instead, we use the ground truth sRGB images to create hard negative and hard positive samples, as detailed below.
\colmap
\subsubsection{Hard negative samples} 
Datasets for sRGB white balance correction include input images rendered with incorrect WB settings, white-balanced ground truth images, and auxiliary images captured at different camera color temperature settings. Specifically, in the Rendered WB dataset Set1 \cite{Afifi_Price_Cohen_Brown_2019} each scene includes a rendering under the correct WB setting (serving as the ground truth), as well as additional images rendered under the following WB settings from the camera: Tungsten (2850 K), Fluorescent (3800 K), Daylight (5500 K), Cloudy (6500 K), and Shade (7500 K). While training, these additional renderings can be used as hard negative samples $I^N_A$ for a given input, $I_A$.
\triplet
\subsubsection{Hard positive samples} A straightforward approach to generate positive samples is geometric augmentations such as random crop or flipping, but these will feature the same underlying scene, creating easily distinguishable samples. In order to encourage the network to learn stronger scene-independent representations, it is essential to generate hard positive samples with different scene content, but with the same WB correction required as the anchor image. Another approach to create positive samples could involve utilizing images from different scenes rendered using the same camera WB setting as the anchor image. However, this approach may not always yield similar samples, as images rendered with the same WB settings might still require different levels of WB correction, as shown in \figref{hard-positive}.

These considerations lead us to the following color transformation technique for generating hard positive samples: given an anchor image $I_A$, its ground truth $I^\text{gt}_A$, along with a separate ground truth image $I^\text{gt}_B$ representing a different scene, we determine a coarse approximation of the WB correction needed to transform $I_{A}$ into the ground truth $I^\text{gt}_A$, and apply the inverse of that transformation to $I^\text{gt}_B$ to obtain the hard positive sample. The generated hard positive sample then needs approximately the same WB correction as $I_A$. This approximation is mediated via the degree 2 polynomial transformation map $\psi:\mathbb{R}^3\to\mathbb{R}^{11}$ defined by $(r,g,b)\mapsto(r,g,b,rg,rb,gb,r^2,b^2,g^2,rgb,1)$ \cite{Afifi_Price_Cohen_Brown_2019}. Specifically, given $I^\text{gt}_A$, we find a $11\times 3$ color correction $M^\star$ satisfying the optimization
\begin{equation}
    M^{\star} := \arg\min_{M} \left\|I_a - M\psi(I^\text{gt}_{A}) \right\|_2
    \label{eq:color-correct}
\end{equation}
representing the inverse of the optimal transformation taking $I_A$ to $I^\text{gt}_A$. Then, given a second ground truth image $I^\text{gt}_B$, we define the positive sample by $I^{P}_A := M^\star\psi(I^\text{gt}_B)$.

\subsection{Architecture details}
Our model has four components, as illustrated in \figref{arch}: a scene classifier, a multi-layer perceptron (MLP), a weight generator, and the LUTs. The scene classifier is composed of 5 convolution layers, each with stride 2 and kernel size 3. The MLP is only used during training to generate anchor, positive, and negative features for contrastive learning. It consists of two linear layers and generates a 128-dimensional feature for each sample. The weight generator, composed of 2 convolution layers, generates $N$ weights corresponding to $N$ basis LUTs.

\quant
\subsection{Loss Functions}
Our loss function combines a standard direct supervision loss measuring pixel-wise error, and a contrastive loss between the anchor $f_A$, positive $f_P$, and negative features $f_N$ from the MLP. Given a batch of $B$ images $\left(I_{A_i}\right)_{i=1}^B$, the direct supervision loss is given by 
\begin{equation}
    L_\text{WB} = \dfrac{1}{B}\sum_{i=1}^B \left\|I^\text{out}_{A_i} - I^\text{gt}_{A_i}\right\|_2,
\end{equation}
where $I^\text{gt}_{A_i}$ is the ground truth image for $I_{A_i}$. The triplet loss involves, for each anchor feature $f_{A_i}$, the corresponding hard positive feature $f^P_{A_i}$ and hard negative feature $f^N_{A_i}$, obtained from corresponding samples as described in \secref{contrastive-learning}:
\begin{equation}
    L_\text{tri} = \dfrac{1}{B}\sum_{i=1}^B \left\|f_{A_i} - f^P_{A_i}\right\|_2 - \left\|f_{A_i} - f^N_{A_i}\right\|_2.
\end{equation}
\ablationqual

We also retain two regularization terms used in \cite{zeng2020learning}: the smooth regularization loss $L_\text{s}$, which encourages smoothness in the values of each 3D LUT across neighbouring indices; and the monotonicity loss $L_\text{m}$, which encourages values of the 3D LUT to increase as the indices increase. 

All together, the model is trained with the following weighted combination of the above loss functions:
\begin{equation}
    L = \lambda_\text{WB}L_\text{WB} + \lambda_\text{tri}L_\text{tri} + \lambda_\text{s}L_\text{s} + \lambda_\text{m}L_\text{m}.
\end{equation}
We choose the following weights empirically:  $\lambda_\text{WB}=1.0$, $\lambda_\text{s}=0.0001$, and $\lambda_\text{m}=10.0$. For the contrastive loss weight, we set $\lambda_\text{tri}=10.0$ for the first 100 epochs and then reduce the effect with $\lambda_\text{tri}=1.0$ for the remainder.

%-------------------------------------------------------------------------

\section{Experiments}
\subsection{Implementation details}
Following prior work \cite{Afifi_Brown_2020, Afifi_Brubaker_Brown_2022, li2023wbflow}, we use 12,000 images from Rendered WB dataset Set1 for training the model. We used Rendered WB dataset Set2 \cite{Afifi_Price_Cohen_Brown_2019} and Rendered Cube \cite{banic2017unsupervised} datasets for our evaluation. Our model, implemented in PyTorch with CUDA, is trained using the Adam optimizer with $\beta_1= 0.9$ and an initial learning rate set to $0.0001$. Training is done on randomly cropped patches of $256\times256$ with random horizontal and vertical flipping. We set batch size as 32 and train all models for 200 epochs. During testing, a downsampled image of size $256\times256$ is used to generate the adaptive LUT, and the look up operation is performed on the high resolution image to achieve real time WB correction. Similar to WBFlow \cite{li2023wbflow}, we report the mean, first (Q1), second (Q2), and third (Q3) quantile of mean angular error (MAE) and  the International Commission on Illumination (CIE) $\Delta\text{E}2000$ measure  for color accuracy \cite{gaurav2005ciede2000}.
\ablationquant
\qual
\subsection{Ablation study}
To illustrate the benefits of the proposed contrastive learning scheme and the novel hard positive mining strategy, we validate various models on Rendered WB Set1 validation split.  Along with RGB color space, we train 3D LUTs in color spaces that separate brightness and color components to emphasize color correction and find that \textit{LAB} color space performs best. These results are presented in \tabref{ablation-quant}. Training with hard positive samples improves mean $\Delta E2000$ by 16\%, compared to 8\% improvement using easy positive samples. Hence, we train WB LUTs in \textit{LAB} color space with hard positive samples. Effects of color temperature on the outputs can be seen in \figref{ablation-qual}, residual blue tint remains in outputs from  3D LUTs, while WB LUTs maintain lower $\Delta\text{E} 2000$ across different color temperatures, showing consistent WB correction across color temperatures.

\subsection{Performance}
Performance of WB LUTs is compared against current state-of-the-art methods in \tabref{quant}. WB LUTs are $300\times$ faster compared to WBFlow and require $12.7\times$ less memory, and fewer parameters compared to any current model. We measure the average run time of all the models on two datasets and include any pre/post-processing necessary to generate outputs at full resolution. 

\subsection {Results}
We compare our method against KNNWB \cite{Afifi_Price_Cohen_Brown_2019}, DeepWB \cite{Afifi_Brown_2020}, MixedWB \cite{Afifi_Brubaker_Brown_2022}, and WBFlow \cite{li2023wbflow}. Each of these methods estimates a lower resolution WB corrected image and uses different post-processing methods like polynomial fitting or bilateral upsampling to generate a high resolution image. 

\tabref{quant} presents performance results of the aforementioned models under MAE and $\Delta E2000$ metrics, on Rendered WB dataset Set2 and the Rendered Cube datasets. Both of these datasets include images taken with different cameras from the training dataset, making them robust tests for generalizability. The results show our model performing competitively against WBFlow under both metrics. Notably, our model shows strong Q2 and Q3 performance, suggesting our contrastive approach succeeds in reigning in poor performance on more extreme input images.

\figref{results} illustrates the effectiveness our method against WB correction from KNNWB \cite{Afifi_Price_Cohen_Brown_2019}, DeepWB \cite{Afifi_Brown_2020}, and MixedWB \cite{Afifi_Brubaker_Brown_2022} models. Our results show comparative WB correction without any artifacts in high resolution outputs generated using available code and their pretrained models. We also see relatively smoother WB correction without large differences in the image colors compared to the ground truth, see KNNWB output in the top row, flowers are colored differently than other methods and the ground truth. Our model consistently generates WB-corrected images of higher quality compared to other models, particularly for images with extreme color temperatures, closely followed by DeepWB, as indicated by the overlaid $\Delta E 2000$ values on the images.

\section{Future work \& Conclusion}
We present WB LUTs, a LUT-based approach for white balance correction improving the quality of baseline 3D LUTs using contrastive learning with hard sample mining. Our model, WB LUTs, achieves near state-of-the-art WB correction on two datasets while generating high resolution images with high efficiency. Although color correction is usually a global image manipulation task, our approach can not handle the cases where a scene is illuminated by two or more illuminants. Future work could include an efficient post-processing step for local color cast correction from different illuminants similar to local contrast enhancement from \cite{zeng2020learning}.

%%%%%%%%% REFERENCES
\bibliographystyle{IEEEbib}
\bibliography{refs}

\end{document}